# Facial Expression Recognition using Deep Learning

Raghu Vamshi N
SCSE
Vellore Institute Of Technology
Chennai,India

Bharathi Raja S
Asisstant Professor(Sr), SCSE
Vellore Institute of Technology
Chennai, India

*Abstract*—Throughout the various ages, facial expressions have become universal ways of non-verbal communication. The ability to recognize facial expressions would pave the path for many novel applications. Regardless of the achievement of customary methodologies in a controlled environment, these methodologies fail on nuanced datasets consisting of partial faces. In this work, I plan to create a deep learning model that can detect people's emotions even under natural conditions. For my work, I take the FER-2013 Dataset. The main goal of this work is to create a system that can detect people's emotions in natural conditions and outperform traditional approaches and human-level performance in terms of accuracy.

*Keywords—Facial expression recognition (FER), Convolution Neural Net (CNN),*

## I. INTRODUCTION

One's emotions play an essential role in their intrapersonal non-verbal communication. However, emotions are abstract and cannot be seen directly with the naked eye. Still, one's emotional state is easily found out by simply observing facial expressions, speech [4] [5], or even text. Of these, facial expressions are one of the most popular due to reasons like; they are visible, and also, it is easier to collect data of facial expressions than for some of the other features. Recently there has been an increase in the need to recognize human emotion due to the development in fields like Human-Computer Interaction, security [5] [8], and testing[6]. Despite knowing facial expression's success under controlled conditions, we have scope for finding emotions in natural situations due to changes in occlusions, pose variations, and illumination.

With the rise of neural networks and other deep learning techniques [7] in the past decade, we achieved remarkable benchmarks for expression recognition under normal conditions, even surpassing the human-level accuracy. This has paved the path to many novel applications in diverse fields like medicine, security, and robotics, which were previously not considered possible.

In this work, my main objective was to understand better and improve the performance of emotion recognition models in the process. I also took some approaches from recent publications, including transfer learning and ensembling, to enhance my model's accuracy. At the same time, I did not use any auxiliary data for my model other than the Dataset to train my models.

## II. RELATED WORKS

In one of the most influential works in the field of emotion recognition by Paul Ekman Wallace [3], six principle emotions were recognized, namely Happy, Sad, Anger, Surprise, Fear, and Disgust. Later on, Neutral was added to the list as a principle emotion, resulting in seven fundamental emotions. So most of the present-day datasets consist of seven emotions, following the initial trend set previously.

Earlier works mostly involved a two-step approach, in which the first step is to extract a few critical features from the images, and in the second step, a classifier (like a neural net or SVM) is used to detect the emotions. This involved hand-crafted features like Histogram of the gradient (HOG)[], Gabor wavelets[], and Haar features. This approach was elegant for simple datasets but not good enough for datasets with a more intra-class variation.

The improvements in deep learning, specifically in the domain of convolution neural nets in computer vision and classification, have led people to develop models to solve the FER problem. One such competition was conducted in Kaggle as a part of the ICML 2013 Workshop on Representation Learning. FER2013 Dataset was introduced in this contest, and most of the traditional approaches weren't able to achieve a reasonable accuracy rate.

The top three teams in the competition all used CNN's in combination with image transformations[]. The winner of the competition Y. Tang used the primal formulation of SVM as a loss function for training and also used the L2-SVM loss function. This gave great results at the time, achieving an accuracy of 71.2% to top the competition.

There is a right amount of existing research in the FER domain; a recent survey paper by S. Li and W.Deng throws light on the present state of deep-learning-based approaches to FER [3].

Also, Shervin minae and Amirali Abdolrashidi[] proposed an end-to-end principle deep learning model to distinguish the emotions, often improving the accuracy relies on increasing the number of layers in the convolution network. Be that as it may, because of the modest amount of classes of feelings, they have proposed a convolution network with under ten layers to accomplish promising outcomes(Got an accuracy of 70.2%).

## III. DATASET

FER is a well-studied field with various datasets available. In this work, we used FER-2013 as our Dataset for both training and testing with no usage of any auxiliary data.

**FER-2013:** This Dataset was first introduced in the ICML 2013 Challenges in Representation learning [1]. Ever since then, this has been used in several other competitions and research papers. This is one of the challenging datasets with human-level accuracy of 65±5%, and the highest performing published works, achieving 75±2%. This Dataset contains 35,887 images of 48x48 resolution, most of which are taken under an uncontrolled setting. FER2013 contains seven facial expressions, with distributions of anger (4953), Disgust (547), Fear (5121), Happy (8989), Sad (6077), Surprise (4002), Neutral (6198).

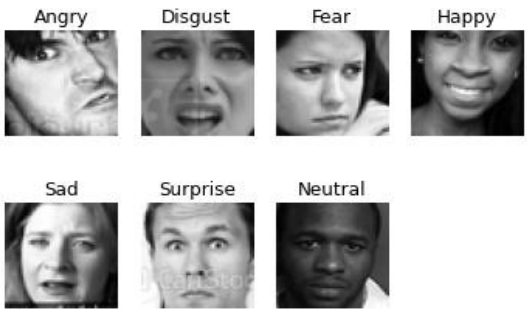

Fig. 1. Images from each class of the FER2013 Dataset

## IV. EXPERIMENTAL SETUP

The dataset is split into training set (28709 images), validation set (3589 images), and testing set (3589 images). All the accuracies we present in the result section is by working on the testing set.

We ran all our models in Google Colabratory storing the dataset and other necessary files in Google drive and mounting it on our colabratory with the basic GPU support.

## V. PROPOSED MODEL

*Baseline Model*

The first model I built consisted of two 3x3x32 same padding convolution layers, two 3x3x64 convolution layers with Relu activations, three 2x2Maxpooling layers, and completed with an FC layer and a softmax layer. Also, batch normalization and 20% dropout layers were included to achieve an accuracy of 61%.

*Five Layer Model*

This is also my model which consists of four stages of two convolution layers in combination with a max-pooling layer, dropout (50%) layers have been added to each convolutional layers block., Fc layer of size 512 and a softmax output layer. In the first convolution layer, L2 (0.01) regularization is used, and in the remaining convolution layers, batch normalization is utilized. All convolution layers apply 32 filters of size 3x3. The max-pooling layers used kernels of size 2x2 and stride 2. "ReLU" has been used as the activation function for all convolutional layers. The categorical cross-entropy function was used as our loss function. I made the loss function to eliminate any "plateaus" by reducing the learning rate parameter of the optimization function with a specific value (0.9 in my case) if the loss function doesn't get better with a particular epoch (3 consecutive epochs). I stop the training process if accuracy does not change for a specific epoch (after eight epochs). With this model, we were able to get an accuracy of 66.2%

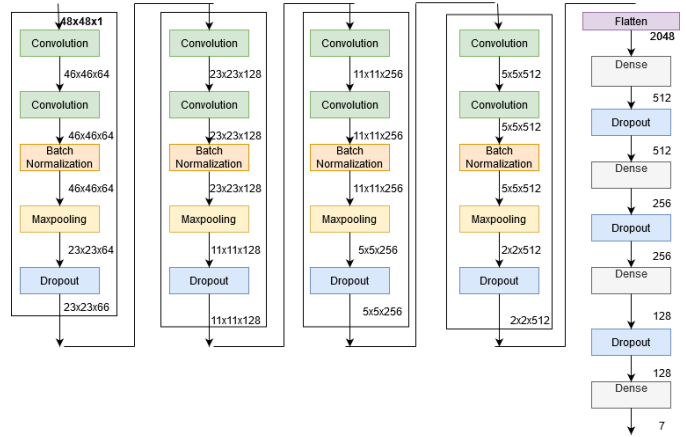

Fig. 2. Example of a figure caption. (*figure caption*)

*Transfer Learning*

Since FER2013 is a relatively small and unbalanced dataset, transfer learning was beneficial for boosting our model's accuracy. I primarily ResNet50 as my pre-trained model from the VGG-face library by freezing some first layers and training on the last few layers to make a prediction. Also, the ResNet50 model had different input requirements than our actual input dimensions. All the images have to be resized from 48x48 to 197x197 aspects and should be colored from grayscale to RGB.

*ResNet50 Model*

This model is a residual network residual net with 50 layers. The Keras implementation of ResNet50 consists of 175 layers similar to the works Brechet [2]. I replaced the output layer with two FC layers (FC 4096 and FC 1024) and a softmax layer that provides a probability for each of the seven emotions. I achieved 73% accuracy on the test set. Freezing the entire pre-trained network came to be unsuccessful since the model failed to fit onto the training set in the initial stages.

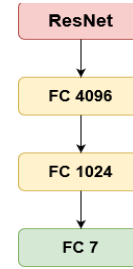

Fig. 3. ResNet50 Module with Transfer Learning.

**ENSEMBLE LEARNING:**

Ensemble learning is performed on the five-layer model and Resnet50 model using soft voting so that both the models are taken into consideration as opposed to hard voting.

**ED NET – EMOTION DETECTION NETWORK:**

This combination of the two models ResNet50 and the five-layer model the novel aspect in our ResNet50 model was that we supplanted the original output layer with FC layers of size 1024 and 4096 and a softmax layer this has given us an accuracy of 73%. Then we further ensemble this with the Five-Layer Model to for our ED NET, the ensembling has improved the accuracy significantly from 73% to 73.92% whose accuracy is comparable to even the state-of-the-art models.

## VI. RESULTS

Table 1 shows the accuracies of some of the models achieved on the FER2013 test data. Results from the Yichuan tang model [1], our baseline, five-layer model, ResNet50[2] model, and the ensemble of the five-layer model and ResNet50 are presented in the table.

TABLE I. RESULTS FOR FER2013 PRIVATE TEST SET

| Model | Depth | Accuracy |
| --- | --- | --- |
| **Human-level** | - | 65±5% |
| **Yichuan Tan [1]** | 4 | 71.2% |
| **Shervin Minaee[5]** | 9 | 70.02% |
| **Baseline** | 5 | 61% |
| **Five-layer model** | 5 | 66.2% |
| **ResNet50** | 50 | 73% |
| **ED NET** | - | 73.9% |
| **State-of-the-art[]** | - | 76.8% |

The state-of-the-art model was an ensemble of six other pre-trained models; most of them from the ImageNet project were fine-tuned and trained using multiple datasets. I cannot recreate such a model due to my limited computing power.

As we see from the figure, there is a flattening of accuracy at 23 epochs.

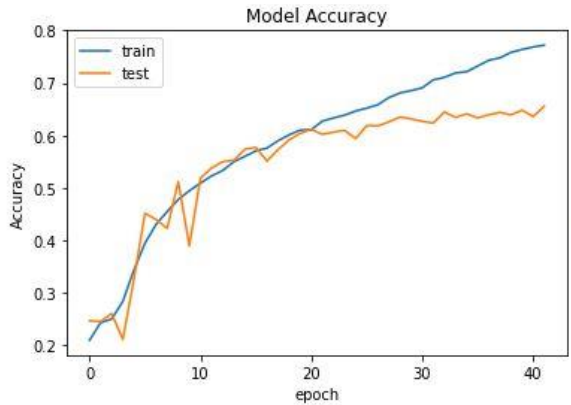

Fig. 4. Model Accuracy for the five-layer model.

The figure shows that there is overfitting of data since we see the test accuracy to flatten just after 25 epochs

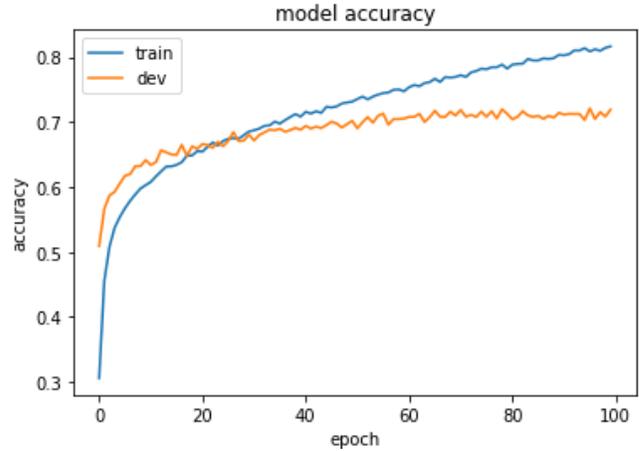

Fig. 5. Model Accuracy for the five-layer model.

## CONCLUSION

The main objective of this paper, in the beginning, was to create a good CNN model to outperform the traditional approaches and human-level accuracy. I was able to achieve my initial objectives with a simple five-layer model. From there on I went through recent publications on using pre-trained models of the ImageNet project for other activities by freezing and manipulating a few layers of the models and found a ResNet50 model and fine-tuned it according to our requirements and ensembles it with our five-layer model to get results that were comparable to even the state-of-the-art models without using any auxiliary data.

## ACKNOWLEDGMENT

I am grateful to Assistant Professor Bharathi Raja S of VIT University, Chennai, India, to guide me in every step of my work. I would like to thank my Program Chair, Dr. Justus S, for giving this brilliant chance.